\documentclass[11pt,a4paper]{article}
\usepackage{iclr2017_conference,times}
\usepackage{url}
\usepackage{amsmath}
\usepackage{times}
\usepackage{latexsym}
\usepackage{multirow}
\usepackage{tikz}
\usepackage{nn}
\usepackage{tikz-dependency}
\mathchardef\mhyphen="2D
\usepackage{url}

\title{Deep Biaffine Attention for Neural\\Dependency Parsing}
\author{
  Timothy Dozat\\
  Stanford University\\
  {\tt tdozat@stanford.edu} \\\And
  Christopher D.\ Manning\\
  Stanford University\\
  {\tt manning@stanford.edu}\\
}
\date{}

\iclrfinalcopy
\begin{document}
\maketitle
\begin{abstract}
  This paper builds off recent work from \citet{KiperwasserGoldberg2016} using neural attention in a simple graph-based dependency parser. We use a larger but more thoroughly regularized parser than other recent BiLSTM-based approaches, with biaffine classifiers to predict arcs and labels. Our parser gets state of the art or near state of the art performance on standard treebanks for six different languages, achieving 95.7\%{} UAS and 94.1\%{} LAS on the most popular English PTB dataset. This makes it the highest-performing graph-based parser on this benchmark---outperforming \citet{KiperwasserGoldberg2016} by 1.8\%{} and 2.2\%{}---and comparable to the highest performing transition-based parser \citep{KuncoroBallesterosetal2016}, which achieves 95.8\%{} UAS and 94.6\%{} LAS. We also show which hyperparameter choices had a significant effect on parsing accuracy, allowing us to achieve large gains over other graph-based approaches.
\end{abstract}

\section{Introduction}
Dependency parsers---which annotate sentences in a way designed to be easy for humans and computers alike to understand---have been found to be extremely useful for a sizable number of NLP tasks, especially those involving natural language understanding in some way \citep{Bowmanetal2016, Angelietal2015, LevyGoldberg2014, Toutanovaetal2016, ParikhPoonToutanova2015}. However, frequent incorrect parses can severely inhibit final performance, so improving the quality of dependency parsers is needed for the improvement and success of these downstream tasks.

The current state-of-the-art transition-based neural dependency parser \citep{KuncoroBallesterosetal2016} substantially outperforms many much simpler neural graph-based parsers. We modify the neural graph-based approach first proposed by \citet{KiperwasserGoldberg2016} in a few ways to achieve competitive performance: we build a network that's larger but uses more regularization; we replace the traditional MLP-based attention mechanism and affine label classifier with biaffine ones; and rather than using the top recurrent states of the LSTM in the biaffine transformations, we first put them through MLP operations that reduce their dimensionality. Furthermore, we compare models trained with different architectures and hyperparameters to motivate our approach empirically. The resulting parser maintains most of the simplicity of neural graph-based approaches while approaching the performance of the SOTA transition-based one.

\section{Background and Related Work}
Transition-based parsers---such as shift-reduce parsers---parse sentences from left to right, maintaining a ``buffer'' of words that have not yet been parsed and a ``stack'' of words whose head has not been seen or whose dependents have not all been fully parsed. At each step, transition-based parsers can access and manipulate the stack and buffer and assign arcs from one word to another. One can then train any multi-class machine learning classifier on features extracted from the stack, buffer, and previous arc actions in order to predict the next action.

\begin{figure}
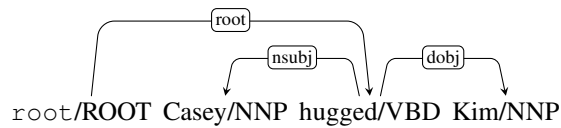

\begin{center}
\begin{dependency}[theme=default]
  \begin{deptext}
    \texttt{root}/ROOT \& Casey/NNP \& hugged/VBD \& Kim/NNP \\
  \end{deptext}
  \depedge{1}{3}{root}
  \depedge{3}{2}{nsubj}
  \depedge{3}{4}{dobj}
\end{dependency}
\end{center}
\caption{A dependency tree parse for \emph{Casey hugged Kim}, including part-of-speech tags and a special \texttt{root} token. Directed edges (or arcs) with labels (or relations) connect the verb to the root and the arguments to the verb head.}
\label{dep}
\end{figure}
 
\citet{ChenManning2014} make the first successful attempt at incorporating deep learning into a transition-based dependency parser. At each step, the (feedforward) network assigns a probability to each action the parser can take based on word, tag, and label embeddings from certain words on the stack and buffer. A number of other researchers have attempted to address some limitations of \citeauthor{ChenManning2014}'s \citeauthor{ChenManning2014} parser by augmenting it with additional complexity: \citet{Weissetal2015} and \citet{Andoretal2016} augment it with a beam search and a conditional random field loss objective to allow the parser to ``undo'' previous actions once it finds evidence that they may have been incorrect; and \citet{Dyeretal2015} and \citep{KuncoroBallesterosetal2016} instead use LSTMs to represent the stack and buffer, getting state-of-the-art performance by building in a way of composing parsed phrases together.

Transition-based parsing processes a sentence sequentially to build up a parse tree one arc at a time. Consequently, these parsers don't use machine learning for directly predicting edges; they use it for predicting the operations of the transition algorithm. Graph-based parsers, by contrast, use machine learning to assign a weight or probability to each possible edge and then construct a maximum spaning tree (MST) from these weighted edges.  \citet{KiperwasserGoldberg2016} present a neural graph-based parser (in addition to a transition-based one) that uses the same kind of attention mechanism as \citet{BahdanauChoBengio2014} for machine translation. In \citeauthor{KiperwasserGoldberg2016}'s \citeyear{KiperwasserGoldberg2016} model, the (bidirectional) LSTM's recurrent output vector for each word is concatenated with each possible head's recurrent vector, and the result is used as input to an MLP that scores each resulting arc. The predicted tree structure at training time is the one where each word depends on its highest-scoring head. Labels are generated analogously, with each word's recurrent output vector and its gold or predicted head word's recurrent vector being used in a multi-class MLP.

Similarly, \citet{Hashimotoetal2016} include a graph-based dependency parser in their multi-task neural model. In addition to training the model with multiple distinct objectives, they replace the traditional MLP-based attention mechanism that \citet{KiperwasserGoldberg2016} use with a bilinear one (but still using an MLP label classifier). This makes it analogous to \citeauthor{Luongetal2015}'s \citeyear{Luongetal2015} proposed attention mechanism for neural machine translation. \citet{Chengetal2016} likewise propose a graph-based neural dependency parser, but in a way that attempts to circumvent the limitation of other neural graph-based parsers being unable to condition the scores of each possible arc on previous parsing decisions. In addition to having one bidirectional recurrent network that computes a recurrent hidden vector for each word, they have additional, unidirectional recurrent networks (left-to-right and right-to-left) that keep track of the probabilities of each previous arc, and use these together to predict the scores for the next arc.

\section{Proposed Dependency Parser}\label{architecture}
\subsection{Deep biaffine attention}
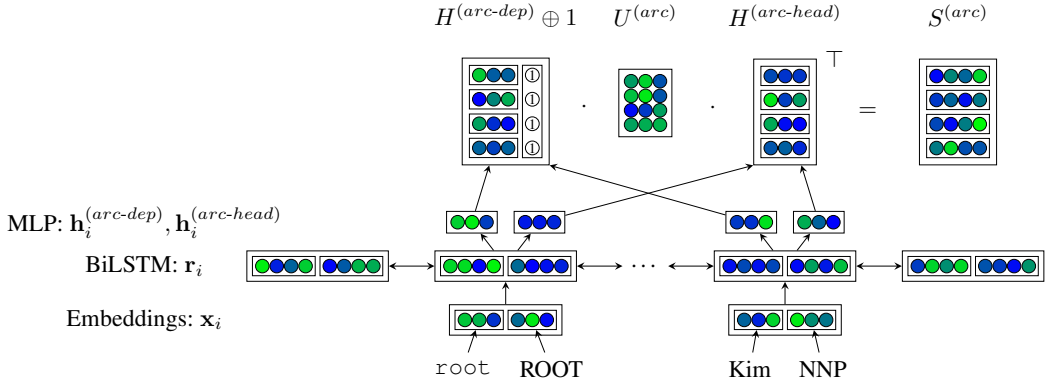
\begin{figure*}
\begin{center}
\resizebox{\textwidth}{!}{
\begin{tikzpicture}[every node/.style={anchor=center}]
  \bivec{r0}{}
  \bivec{r1}{right=.67cm of r0}
  \node[right=.67cm of r1] (dots) {\ldots};
  \bivec{rn}{right=.67cm of dots}
  \bivec{rnp1}{right=.67cm of rn}
  
  \wordvec{x1}{below=.33cm of r1}
  \wordvec{xn}{below=.33cm of rn}
  
  \mlpvec{m1-1}{above=.33cm of r1-1}
  \mlpvec{m1-2}{above=.33cm of r1-2}
  \mlpvec{mn-1}{above=.33cm of rn-1}
  \mlpvec{mn-2}{above=.33cm of rn-2}
  
  \node[below left=.33cm and -.67cm of x1-1] (w1) {\texttt{root}};
  \node[below right=.33cm and -.67cm of x1-2] (t1) {ROOT};
  \node[below left=.33cm and -.67cm of xn-1] (wn) {Kim};
  \node[below right=.33cm and -.67cm of xn-2] (tn) {NNP};
  
  \catstack{dep}{above=1.25cm of r1}
  \matstack{head}{above=1.25cm of rn}
  \tensor{ten}{above=1.75cm of dots}
  \node[right=0cm of head.north east] {$\top$};
  \bigstack{scores}{right=1.5cm of head}
  \path (dep) -- node{$\cdot$} (ten);
  \path (ten) -- node{$\cdot$} (head);
  \path (head) -- node{$=$} (scores);
  
  \node[left=.5cm of r0] (lstm) {BiLSTM: $\mathbf{r}_i$};
  \node at (x1 -| lstm) {Embeddings: $\mathbf{x}_i$};
  \node at (m1-1 -| lstm) {MLP: $\mathbf{h}^{(arc\mhyphen{}dep)}_i,\mathbf{h}^{(arc\mhyphen{}head)}_i$};
  \node[above=.33cm of dep] (H) {$H^{(arc\mhyphen{}dep)}\oplus 1$};
  \node at (H -| ten) {$U^{(arc)}$};
  \node at (H -| head) {$H^{(arc\mhyphen{}head)}$};
  \node at (H -| scores) {$S^{(arc)}$};
  
  \draw[<->, >=stealth] (r0) -- (r1);
  \draw[<->, >=stealth] (r1) -- (dots);
  \draw[<->, >=stealth] (dots) -- (rn);
  \draw[<->, >=stealth] (rn) -- (rnp1);
  
  \draw[->, >=stealth] (x1) -- (r1);
  \draw[->, >=stealth] (xn) -- (rn);
  
  \draw[->, >=stealth] (w1) -- (x1-1);
  \draw[->, >=stealth] (t1) -- (x1-2);
  \draw[->, >=stealth] (wn) -- (xn-1);
  \draw[->, >=stealth] (tn) -- (xn-2);
  
  \draw[->, >=stealth] (r1) -- (m1-1);
  \draw[->, >=stealth] (r1) -- (m1-2);
  \draw[->, >=stealth] (rn) -- (mn-1);
  \draw[->, >=stealth] (rn) -- (mn-2);
  
  \draw[->, >=stealth] (m1-1) -- (dep);
  \draw[->, >=stealth] (mn-1) -- (dep.south east);
  \draw[->, >=stealth] (m1-2) -- (head.south west);
  \draw[->, >=stealth] (mn-2) -- (head);
\end{tikzpicture}}
\end{center}
\caption{BiLSTM with deep biaffine attention to score each possible head for each dependent, applied to the sentence ``Casey hugged Kim''. We reverse the order of the biaffine transformation here for clarity.}
\label{proposed model}
\end{figure*}

We make a few modifications to the graph-based architectures of \citet{KiperwasserGoldberg2016}, \citet{Hashimotoetal2016}, and \citet{Chengetal2016}, shown in Figure \ref{proposed model}: we use biaffine attention instead of bilinear or traditional MLP-based attention; we use a biaffine dependency label classifier; and we apply dimension-reducing MLPs to each recurrent output vector $\mathbf{r}_i$ before applying the biaffine transformation.\footnote{In this paper we follow the convention of using lowercase italic letters for scalars and indices, lowercase bold letters for vectors, uppercase italic letters for matrices, uppercase bold letters for higher order tensors. We also maintain this notation when indexing; so row $i$ of matrix $R$ would be represented as $\mathbf{r}_i$.} The choice of biaffine rather than bilinear or MLP mechanisms makes the classifiers in our model analogous to traditional affine classifiers, which use an affine transformation over a single LSTM output state $\mathbf{r}_i$ (or other vector input) to predict the vector of scores $\mathbf{s}_i$ for all classes (\ref{tra-class}). We can think of the proposed biaffine attention mechanism as being a traditional affine classifier, but using a $(d \times d)$ linear transformation of the stacked LSTM output $RU^{(1)}$ in place of the weight matrix $W$ and a $(d \times 1)$ transformation $R\mathbf{u}^{(2)}$ for the bias term $\mathbf{b}$ (\ref{biaff-att}). 
\begin{align}
  \label{tra-class}\mathbf{s}_i &= W\mathbf{r}_i + \mathbf{b} & \text{Fixed-class affine classifier}\\
  \label{biaff-att}\mathbf{s}_i^{(arc)} &= \left(RU^{(1)}\right)\mathbf{r}_i + \left(R\mathbf{u}^{(2)}\right) & \text{Variable-class biaffine classifier}
\end{align}
In addition to being arguably simpler than the MLP-based approach (involving one bilinear layer rather than two linear layers and a nonlinearity), this has the conceptual advantage of directly modeling both the prior probability of a word $j$ receiving any dependents in the term $\mathbf{r}_j^\top\mathbf{u}^{(2)}$ and the likelihood of $j$ receiving a specific dependent $i$ in the term $\mathbf{r}^\top_jU^{(1)}\mathbf{r}_i$. Analogously, we also use a biaffine classifier to predict dependency labels given the gold or predicted head $y_i$ (\ref{biaff-class}).
\begin{align}
  \label{biaff-class}\mathbf{s}_i^{(label)} &= \mathbf{r}_{y_i}^\top \mathbf{U}^{(1)}\mathbf{r}_i + (\mathbf{r}_{y_i} \oplus \mathbf{r}_i)^\top U^{(2)} + \mathbf{b} & \text{Fixed-class biaffine classifier}
\end{align}
This likewise directly models each of the prior probability of each class, the likelihood of a class given just word $i$ (how probable a word is to take a particular label), the likelihood of a class given just the head word $y_i$ (how probable a word is to take dependents with a particular label), and the likelihood of a class given both word $i$ and its head (how probable a word is to take a particular label given that word's head). 
 
Applying smaller MLPs to the recurrent output states before the biaffine classifier has the advantage of stripping away information not relevant to the current decision. That is, every top recurrent state $\mathbf{r}_i$ will need to carry enough information to identify word $i$'s head, find all its dependents, exclude all its non-dependents, assign itself the correct label, and assign all its dependents their correct labels, as well as transfer any relevant information to the recurrent states of words before and after it. Thus $\mathbf{r}_i$ necessarily contains significantly more information than is needed to compute any individual score, and training on this superfluous information needlessly reduces parsing speed and increases the risk of overfitting. Reducing dimensionality and applying a nonlinearity (\ref{arc-dep} - \ref{arc-score}) addresses both of these problems. We call this a \emph{deep} bilinear attention mechanism, as opposed to \emph{shallow} bilinear attention, which uses the recurrent states directly.
\begin{align}
  \label{arc-dep}\mathbf{h}_i^{(arc\mhyphen{}dep)} &= \text{MLP}^{(arc\mhyphen{}dep)}(\mathbf{r}_i)\\
  \mathbf{h}_j^{(arc\mhyphen{}head)} &= \text{MLP}^{(arc\mhyphen{}head)}(\mathbf{r}_j)\\
  \label{arc-score}\mathbf{s}^{(arc)}_i &= H^{(arc\mhyphen{}head)}U^{(1)}\mathbf{h}^{(arc\mhyphen{}dep)}_i\\ 
  &\nonumber\quad+ H^{(arc\mhyphen{}head)}\mathbf{u}^{(2)}
\end{align}

We apply MLPs to the recurrent states before using them in the label classifier as well. As with other graph-based models, the predicted tree at training time is the one where each word is a dependent of its highest scoring head (although at test time we ensure that the parse is a well-formed tree via the MST algorithm).

\subsection{Hyperparameter configuration}
\begin{table}[ht]
  \begin{center}\small
  \begin{tabular}{llll}
    \bf Param & \bf Value & \bf Param & \bf Value\\
    Embedding size & 100 & Embedding dropout & 33\%\\
    LSTM size & 400 & LSTM dropout & 33\%\\
    Arc MLP size & 500 & Arc MLP dropout & 33\%\\
    Label MLP size & 100 & Label MLP dropout & 33\%\\
    LSTM depth & 3 & MLP depth & 1\\
    $\alpha$ & $2e^{-3}$ & $\beta_1$,$\beta_2$ & .9 \\
    Annealing & $.75^{\frac{t}{5000}}$ & $t_{max}$ & 50,000
  \end{tabular}
  \caption{Model hyperparameters}
  \label{Hyperparams}
  \end{center}
\end{table}

Aside from architectural differences between ours and the other graph-based parsers, we make a number of hyperparameter choices that allow us to outperform theirs, laid out in Table \ref{Hyperparams}. We use 100-dimensional uncased word vectors\footnote{We compute a ``trained'' embedding matrix composed of words that occur at least twice in the training dataset and add these embeddings to their corresponding pretrained embeddings. Any words that don't occur in either embedding matrix are replaced with a separate OOV token.} and POS tag vectors; three BiLSTM layers (400 dimensions in each direction); and 500- and 100-dimensional ReLU MLP layers. We also apply dropout at every stage of the model: we drop words and tags (independently); we drop nodes in the LSTM layers (input and recurrent connections), applying the same dropout mask at every recurrent timestep (cf.\ the Bayesian dropout of \citet{GalGhaharmani2015}); and we drop nodes in the MLP layers and classifiers, likewise applying the same dropout mask at every timestep. We optimize the network with annealed Adam \citep{KingmaBa2014} for about 50,000 steps, rounded up to the nearest epoch.

\section{Experiments \&{} Results}
\subsection{Datasets}
We show test results for the proposed model on the English Penn Treebank, converted into Stanford Dependencies using both version 3.3.0 and version 3.5.0 of the Stanford Dependency converter (PTB-SD 3.3.0 and PTB-SD 3.5.0); the Chinese Penn Treebank; and the CoNLL 09 shared task dataset,\footnote{We exclude the Japanese dataset from our evaluation because we do not have access to it.} following standard practices for each dataset. We omit punctuation from evaluation only for the PTB-SD and CTB. For the English PTB-SD datasets, we use POS tags generated from the Stanford POS tagger \citep{Toutanovaetal2003}; for the Chinese PTB dataset we use gold tags; and for the CoNLL 09 dataset we use the provided predicted tags. Our hyperparameter search was done with the PTB-SD 3.5.0 validation dataset in order to minimize overfitting to the more popular PTB-SD 3.3.0 benchmark, and in our hyperparameter analysis in the following section we report performance on the PTB-SD 3.5.0 test set, shown in Tables \ref{Dev results 1} and \ref{Dev results 2}.

\subsection{Hyperparameter choices}
\begin{table}
  \begin{center}\small
  \begin{tabular}{llll@{\hskip 2em}lllllll}
    \multicolumn{4}{c}{\bf Classifier} & \multicolumn{4}{c}{\bf Size}\\
    \bf Model & \bf UAS & \bf LAS & \bf Sents/sec  & \bf Model & \bf UAS & \bf LAS & \bf Sents/sec \\
    Deep & \bf 95.75 & \bf 94.22 & \bf 410.91      & 3 layers, 400d & 95.75 & 94.22 & 410.91       \\
    Shallow & 95.74 & 94.00* & 298.99              & 3 layers, 300d & 95.82 & \bf 94.24 & 460.01   \\
    Shallow, 50\%{} drop & 95.73 & 94.05* & 300.04 & 3 layers, 200d & 95.55* & 93.89* & 469.45     \\
    Shallow, 300d & 95.63* & 93.86* & 373.24       & 2 layers, 400d & 95.62* & 93.98* & \bf 497.99 \\
    MLP & 95.53* & 93.91* & 367.44                 & 4 layers, 400d & \bf 95.83 & 94.22 & 362.09      \\[2ex]
    \multicolumn{4}{c}{\bf Recurrent Cell}\\
    \bf Model & \bf UAS & \bf LAS & \bf Sents/sec\\
    LSTM & \bf 95.75 & \bf 94.22 & 410.91\\
    GRU & 93.18* & 91.08* & 435.32\\
    Cif-LSTM & 95.67 & 94.06* & \bf 463.25\\
  \end{tabular}
  \end{center}
  \caption{Test accuracy and speed on PTB-SD 3.5.0. Statistically significant differences are marked with an asterisk.}
  \label{Dev results 1}
\end{table}
\begin{table}
  \begin{center}\small
  \begin{tabular}{lll@{\hskip 2em}lll}
    \multicolumn{3}{c}{\bf Input Dropout} & \multicolumn{3}{c}{\bf Adam}\\
    \bf Model & \bf UAS & \bf LAS    & \bf Model & \bf UAS & \bf LAS\\       
    Default & \bf 95.75 & \bf 94.22  & $\beta_2=.9$ & \bf 95.75 & \bf 94.22 \\
    No word dropout & 95.74 & 94.08* & $\beta_2=.999$ & 95.53* & 93.91*\\
    No tag dropout & 95.28* & 93.60* \\
    No tags & 95.77 & 93.91*         
  \end{tabular}
  \end{center}
  \caption{Test Accuracy on PTB-SD 3.5.0. Statistically significant differences are marked with an asterisk.}
  \label{Dev results 2}
\end{table}
\subsubsection{Attention mechanism}
We examined the effect of different classifier architectures on accuracy and performance. What we see is that the deep bilinear model outperforms the others with respect to both speed and accuracy. The model with shallow bilinear arc and label classifiers gets the same unlabeled performance as the deep model with the same settings, but because the label classifier is much larger ($(801 \times c \times 801)$ as opposed to $(101 \times c \times 101)$), it runs much slower and overfits. One way to decrease this overfitting is by increasing the MLP dropout, but that of course doesn't change parsing speed; another way is to decrease the recurrent size to 300, but this hinders unlabeled accuracy without increasing parsing speed up to the same levels as our deeper model. We also implemented the MLP-based approach to attention and classification used in \citet{KiperwasserGoldberg2016}.\footnote{In the version of TensorFlow we used, the model's memory requirements during training exceeded the available memory on a single GPU when default settings were used, so we reduced the MLP hidden size to 200} We found this version to likewise be somewhat slower and significantly underperform the deep biaffine approach in both labeled and unlabeled accuracy.

\subsubsection{Network size}
We also examine more closely how network size influences speed and accuracy. In \citeauthor{KiperwasserGoldberg2016}'s \citeyear{KiperwasserGoldberg2016} model, the network uses 2 layers of 125-dimensional bidirectional LSTMs; in \citeauthor{Hashimotoetal2016}'s \citeyear{Hashimotoetal2016} model, it has one layer of 100-dimensional bidirectional LSTMs dedicated to parsing (two lower layers are also trained on other objectives); and \citeauthor{Chengetal2016}'s \citeyear{Chengetal2016} model has one layer of 368-dimensional GRU cells. We find that using three or four layers gets significantly better performance than two layers, and increasing the LSTM sizes from 200 to 300 or 400 dimensions likewise signficantly improves performance.\footnote{The model with 400-dimensional recurrent states significantly outperforms the 300-dimensional one on the validation set, but not on the test set}

\subsubsection{Recurrent cell}
GRU cells have been promoted as a faster and simpler alternative to LSTM cells, and are used in the approach of \citet{Chengetal2016}; however, in our model they drastically underperformed LSTM cells. We also implemented the coupled input-forget gate LSTM cells (Cif-LSTM) suggested by \citet{Greffetal2015},\footnote{In addition to using a coupled input-forget gate, we remove the first tanh nonlinearity, which is no longer needed when using a coupled gate} finding that while the resulting model still slightly underperforms the more popular LSTM cells, the difference between the two is much smaller. Additionally, because the gate and candidate cell activations can be computed simultaneously with one matrix multiplication, the Cif-LSTM model is faster than the GRU version even though they have the same number of parameters. We hypothesize that the output gate in the Cif-LSTM model allows it to maintain a sparse recurrent output state, which helps it adapt to the high levels of dropout needed to prevent overfitting in a way that GRU cells are unable to do. 

\subsubsection{Embedding Dropout}
Because we increase the parser's power, we also have to increase its regularization. In addition to using relatively extreme dropout in the recurrent and MLP layers mentioned in Table \ref{Hyperparams}, we also regularize the input layer. We drop 33\%{} of words and 33\%{} of tags during training: when one is dropped the other is scaled by a factor of two to compensate, and when both are dropped together, the model simply gets an input of zeros. Models trained with only word or tag dropout but not both wind up signficantly overfitting, hindering label accuracy and---in the latter case---attachment accuracy. Interestingly, not using any tags at all actually results in better performance than using tags without dropout.

\subsubsection{Optimizer}
We choose to optimize with Adam \citep{KingmaBa2014}, which (among other things) keeps a moving average of the $L_2$ norm of the gradient for each parameter throughout training and divides the gradient for each parameter by this moving average, ensuring that the magnitude of the gradients will on average be close to one. However, we find that the value for $\beta_2$ recommended by \citeauthor{KingmaBa2014}---which controls the decay rate for this moving average---is too high for this task (and we suspect more generally). When this value is very large, the magnitude of the current update is heavily influenced by the larger magnitude of gradients very far in the past, with the effect that the optimizer can't adapt quickly to recent changes in the model. Thus we find that setting $\beta_2$ to $.9$ instead of $.999$ makes a large positive impact on final performance.

\begin{table}
  \begin{center}\small
  \begin{tabular}{llllll}
    && \multicolumn{2}{c}{\bf English PTB-SD 3.3.0} & \multicolumn{2}{c}{\bf Chinese PTB 5.1}\\
    \bf Type & \bf Model & \bf UAS & \bf LAS & \bf UAS & \bf LAS\\[1ex]
    \multirow{3}{*}{Transition} & \citet{Ballesterosetal2016} & 93.56 & 91.42  & 87.65 & 86.21\\
    &\citet{Andoretal2016} & 94.61 & 92.79 & -- & --\\
    &\citet{KuncoroBallesterosetal2016} &\bf 95.8 & \bf 94.6 & -- & --\\[1ex]
    \multirow{4}{*}{Graph} & \citet{KiperwasserGoldberg2016} & 93.9 & 91.9 & 87.6 & 86.1\\
    &\citet{Chengetal2016} & 94.10 & 91.49 & 88.1 & 85.7\\
    &\citet{Hashimotoetal2016} & 94.67 & 92.90 & -- & --\\
    &Deep Biaffine & 95.74 & 94.08 & \bf 89.30 & \bf 88.23
  \end{tabular}
  \end{center}
  \caption{Results on the English PTB and Chinese PTB parsing datasets}
  \label{compare}
\end{table}
\begin{table}
  \begin{center}\small
  \begin{tabular}{lllllll}
    & \multicolumn{2}{c}{\bf Catalan} & \multicolumn{2}{c}{\bf Chinese} & \multicolumn{2}{c}{\bf Czech}\\
    \bf Model & \bf UAS & \bf LAS & \bf UAS & \bf LAS & \bf UAS & \bf LAS \\[1ex]
    \citeauthor{Andoretal2016} & 92.67 & 89.83 & 84.72 & 80.85 & 88.94 & 84.56\\
    Deep Biaffine & \bf 94.69 & \bf 92.02 & \bf 88.90 & \bf 85.38 & \bf 92.08 & \bf 87.38\\[2ex]
    & \multicolumn{2}{c}{\bf English} & \multicolumn{2}{c}{\bf German}& \multicolumn{2}{c}{\bf Spanish}\\
    \bf Model & \bf UAS & \bf LAS & \bf UAS & \bf LAS & \bf UAS & \bf LAS\\[1ex]
    \citeauthor{Andoretal2016} & 93.22 & 91.23 & 90.91 & 89.15 & 92.62 & 89.95\\
    Deep Biaffine & \bf 95.21 & \bf 93.20 & \bf 93.46 & \bf 91.44 & \bf 94.34 & \bf 91.65
  \end{tabular}
  \caption{Results on the CoNLL '09 shared task datasets}
  \end{center}
\end{table}

\subsection{Results}
Our model gets nearly the same UAS performance on PTB-SD 3.3.0 as the current SOTA model from \citet{KuncoroBallesterosetal2016} in spite of its substantially simpler architecture, and gets SOTA UAS performance on CTB 5.1\footnote{We'd like to thank Zhiyang Teng for finding a bug in the original code that affected the CTB 5.1 dataset} as well as SOTA performance on all CoNLL 09 languages. It is worth noting that the CoNLL 09 datasets contain many non-projective dependencies, which are difficult or impossible for transition-based---but not graph-based---parsers to predict. This may account for some of the large, consistent difference between our model and \citeauthor{Andoretal2016}'s \citeyear{Andoretal2016} transition-based model applied to these datasets.

Where our model appears to lag behind the SOTA model is in LAS, indicating one of a few possibilities. Firstly, it may be the result of inefficiencies or errors in the GloVe embeddings or POS tagger, in which case using alternative pretrained embeddings or a more accurate tagger might improve label classification. Secondly, the SOTA model is specifically designed to capture phrasal compositionality; so another possibility is that ours doesn't capture this compositionality as effectively, and that this results in a worse label score. Similarly, it may be the result of a more general limitation of graph-based parsers, which have access to less explicit syntactic information than transition-based parsers when making decisions. Addressing these latter two limitations would require a more innovative architecture than the relatively simple one used in current neural graph-based parsers.

\section{Conclusion}
In this paper we proposed using a modified version of bilinear attention in a neural dependency parser that increases parsing speed without hurting performance. We showed that our larger but more regularized network outperforms other neural graph-based parsers and gets comparable performance to the current SOTA transition-based parser. We also provided empirical motivation for the proposed architecture and configuration over similar ones in the existing literature. Future work will involve exploring ways of bridging the gap between labeled and unlabeled accuracy and augment the parser with a smarter way of handling out-of-vocabulary tokens for morphologically richer languages.

\bibliography{tdozat}
\end{document}